\documentclass[conference]{IEEEtran}
\usepackage{xcolor}
\usepackage{url}
\usepackage{acronym}
\usepackage{multirow}
\usepackage{adjustbox}
\usepackage{graphicx}
\usepackage{amsmath}
\usepackage{subcaption}
\usepackage{filecontents}
\usepackage{colortbl}
\usepackage{textcomp} % for arrow in text
\usepackage{enumitem}
\usepackage[ruled]{algorithm2e}
\usepackage{threeparttable}
\usepackage{tablefootnote}
\usepackage{lipsum}
\usepackage{todonotes}
\usepackage[mathlines,switch]{lineno}
\usepackage{pifont}% http://ctan.org/pkg/pifont
\usepackage[english]{babel}
\usepackage{tikz}
\usepackage{setspace}
\usepackage{booktabs}
\usetikzlibrary{positioning}
\usepackage{cite}

\begin{document}

\title{
% Autonomous Exploration Agent Driven Automated Anomaly Detection in Games
%Autonomous Exploration Agents for Visual Anomaly Detection in Video Games
Evaluating VLMs for Autonomous Agent-Driven Geometry Clipping Detection in Video Game QA
} % looks good.

%\author{Benedict Wilkins, Carlos Celemin, Adrián Barahona-Ríos, Saman Zadtootaghaj and Nabajeet Barman \\
%Sony Interactive Entertainment, London, United Kingdom}

\author{Carlos Celemin*\thanks{*All authors contributed equally to this work.}, Benedict Wilkins*, Adrián Barahona-Ríos*, Saman Zadtootaghaj* and Nabajeet Barman* \\
Sony Interactive Entertainment, London, United Kingdom\\
\{carlos.celemin, benedict.wilkins, adrian.barahona.rios, saman.zadtootaghaj, nabajeet.barman\}@sony.com
}

\IEEEoverridecommandlockouts

\maketitle
\begin{abstract}

% LATEST VERSION BEFORE REVISION BELOW IT
% In this work, we study the use of Vision-Language Models (VLMs) for anomaly detection in an agent-driven game Quality Assurance (QA) pipeline. 
% As a case study, we focus on geometry clipping. 
% In this evaluation, an exploration agent navigates a game level to collect visual observations, while our implemented automatic annotation pipeline provides frame-level clipping labels. 
% This setup allows us to evaluate recent VLMs on a controlled anomaly detection task without manual annotation.

% Our results show that VLMs can capture visual cues associated with geometry clipping, but their predictions remain noisy, with a substantial number of false positives. Qualitative inspection suggests that many of these false positives are not arbitrary errors, but visually ambiguous cases involving occlusions, suspicious geometry or near-clipping cases.

% Overall, this work provides an initial study of VLMs for geometry clipping detection in agent-driven game QA, highlighting both their potential as candidate filters and their limitations under frame-level evaluation.

In this work, we study the use of Vision-Language Models (VLMs) for anomaly detection in an agent-driven game Quality Assurance (QA) pipeline focusing on geometry clipping. In this evaluation, a custom exploration agent navigates a game level to collect visual observations, while the automatic annotation pipeline provides frame-level clipping labels. This setup allows us to evaluate recent VLMs on a controlled anomaly detection task without manual annotation. We benchmark six recent VLMs (Gemini, GPT, Qwen, Gemma, Llama, and Ministral) under a zero-shot prompting setting and analyse their sensitivity to four prompt variants.

Our results show that while the VLMs can capture visual cues associated with geometry clipping, they all produce substantial false positives on visually ambiguous frames such as near-contact geometry and partial occlusions. Gemini-3.1-Flash achieves the best overall accuracy and is the most robust to prompt variation, while open-source models exhibit large precision--recall swings depending on the prompt design. These findings suggest that current VLMs are best suited as high-recall candidate filters within multi-stage QA pipelines rather than as standalone bug detectors.

\end{abstract}


\begin{tikzpicture}[overlay, remember picture]
\path (current page.north) node (anchor) {};
\node [below=of anchor] {%
IEEE Conference on Games 2026};
\end{tikzpicture}

\IEEEpeerreviewmaketitle

\section{Introduction}\label{sec:intro}

% BEN -- i modified this to be more suited to the new setup

QA is one of the most critical problems of modern video game development, ensuring that complex interactive systems behave as intended across diverse scenarios. However, as games grow in scale and complexity, exhaustive manual testing becomes increasingly difficult and time-consuming. 
They feature a combinatorial explosion of possible states, making it challenging to detect all potential issues prior to release. As a result, visual anomalies (e.g., rendering artifacts, missing objects, or clipping/out of bounds) often persist into shipped products, negatively impacting user experience, and increasing post-release maintenance costs. To address these scalability challenges, game playing agents are actively being developed to explore game environments with minimal human intervention \cite{lu2022go,tufano2022using, spick2024behaviouralcloningvizdoom, amadori2024robust, celemin2025learning}. 
These agents can traverse large state spaces and enable broader test coverage than manual QA through scalable or targeted exploration, but they do not by themselves address a core aspect of QA: detecting bugs.

Existing learning-based approaches for video game bug detection have explored temporal anomaly detection from frame sequences and supervised models trained on labelled or synthetic examples \cite{azizi2024astrobug, paduraru2026state}. 
However, these approaches often depend on game-specific data or small-scale supervised datasets, which may limit their ability to generalise across diverse visual styles. 
Moreover, because obtaining game-specific bug labels is challenging, supervised approaches may be difficult to apply in practice. Recent advances in Vision-Language Models (VLMs) offer a promising alternative by combining visual understanding with natural language reasoning \cite{zhang2024vision}. 
VLMs learn from massive training datasets and make use of inference-time context, which enables generalization across diverse settings. 
Although they show promise for automated game QA, recent VLM-focused benchmarks indicate that current models struggle to detect certain classes of bugs ~\cite{cao2024physgame, taesiri2025videogameqa,yakun2026eccv}. 
Despite recent efforts, it is also not yet clear how best to utilize their existing capabilities for practical game QA.

\begin{figure}[t]
\centering
\begin{subfigure}{0.99\columnwidth}
\centering
\includegraphics[width=0.495\columnwidth]{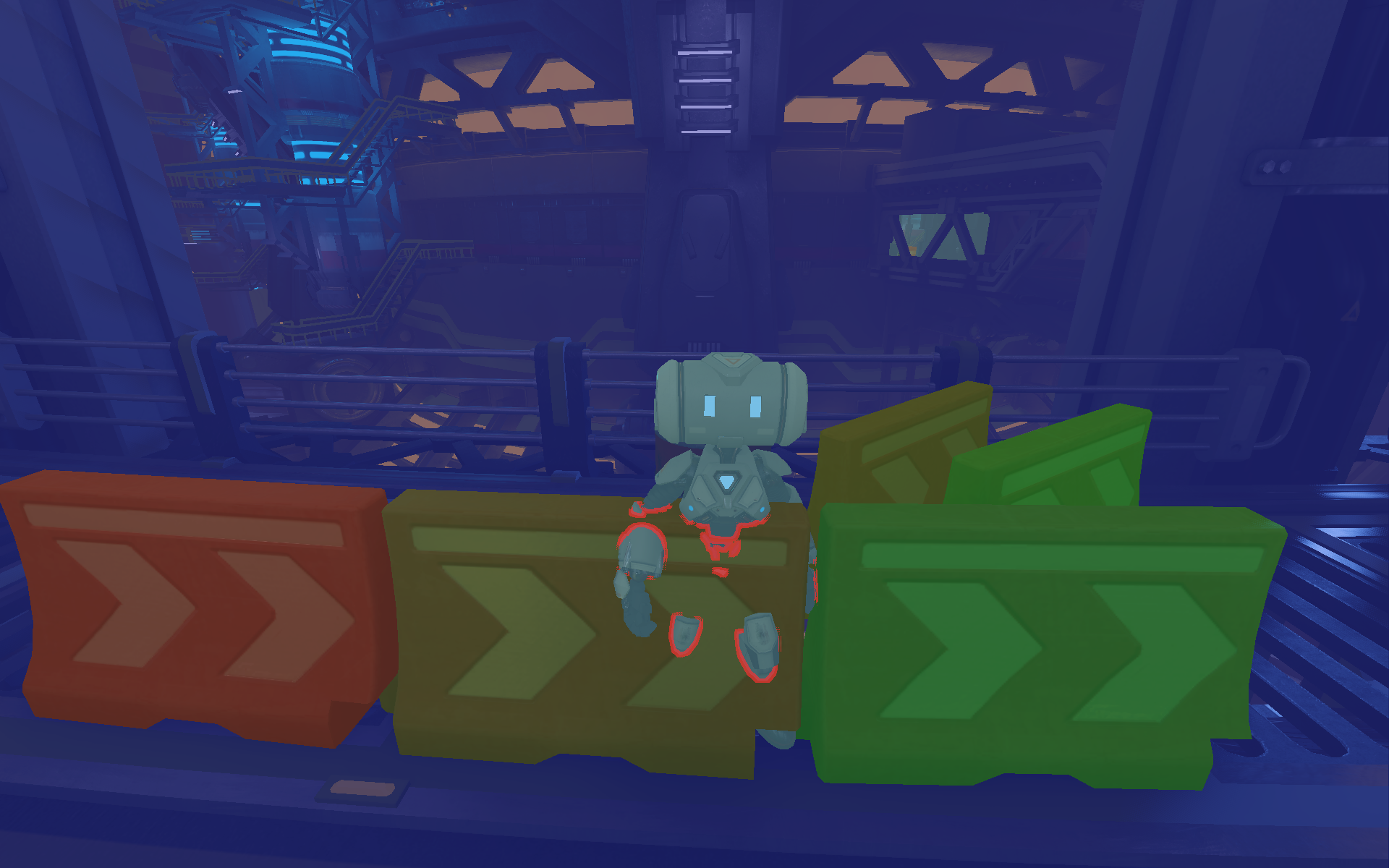}\hfill\includegraphics[width=0.495\columnwidth]{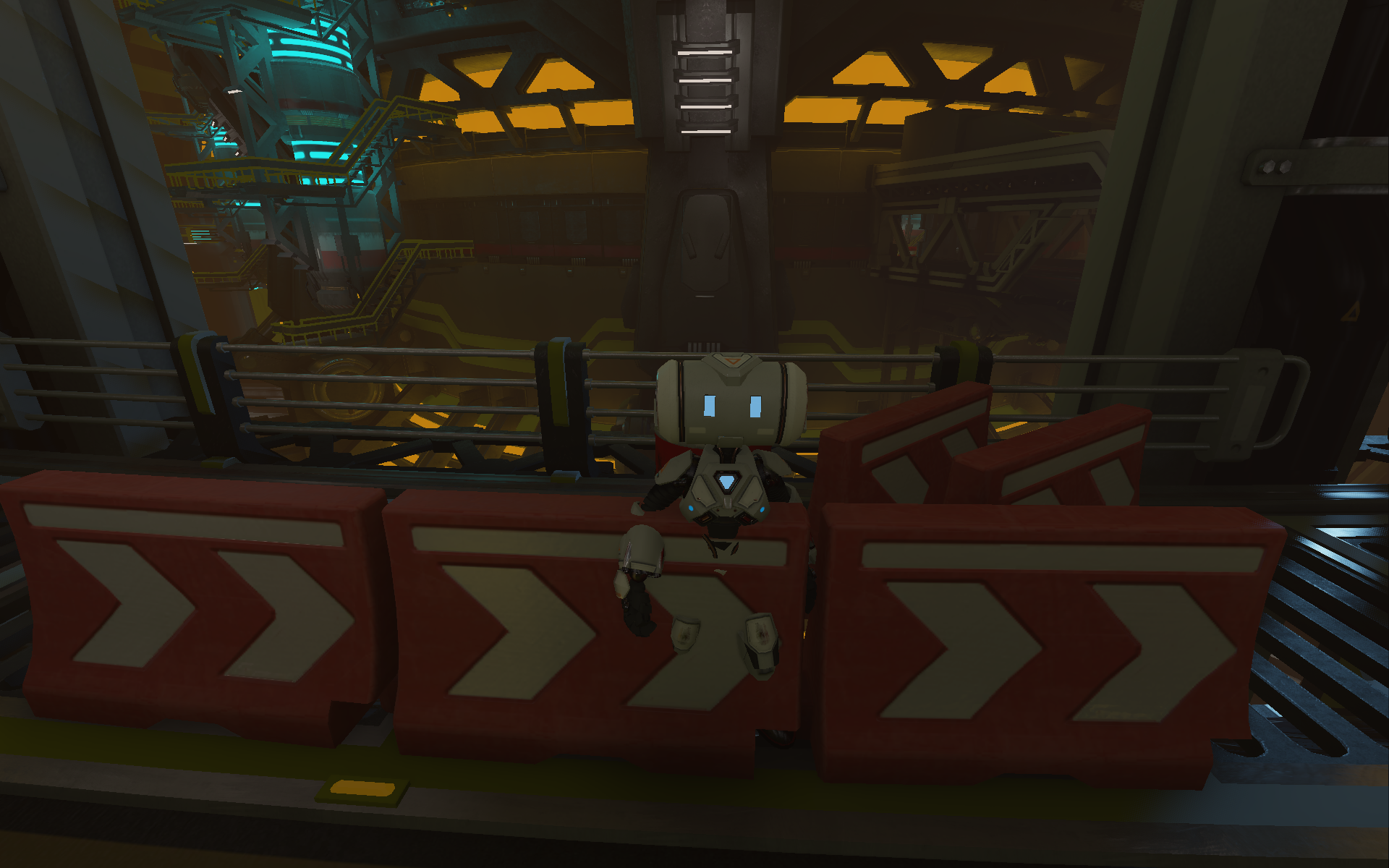}
%\caption{Easy Clipping Bug}
\end{subfigure}

\vspace{0.2em}

\begin{subfigure}{0.99\columnwidth}
\centering
\includegraphics[width=0.495\columnwidth]{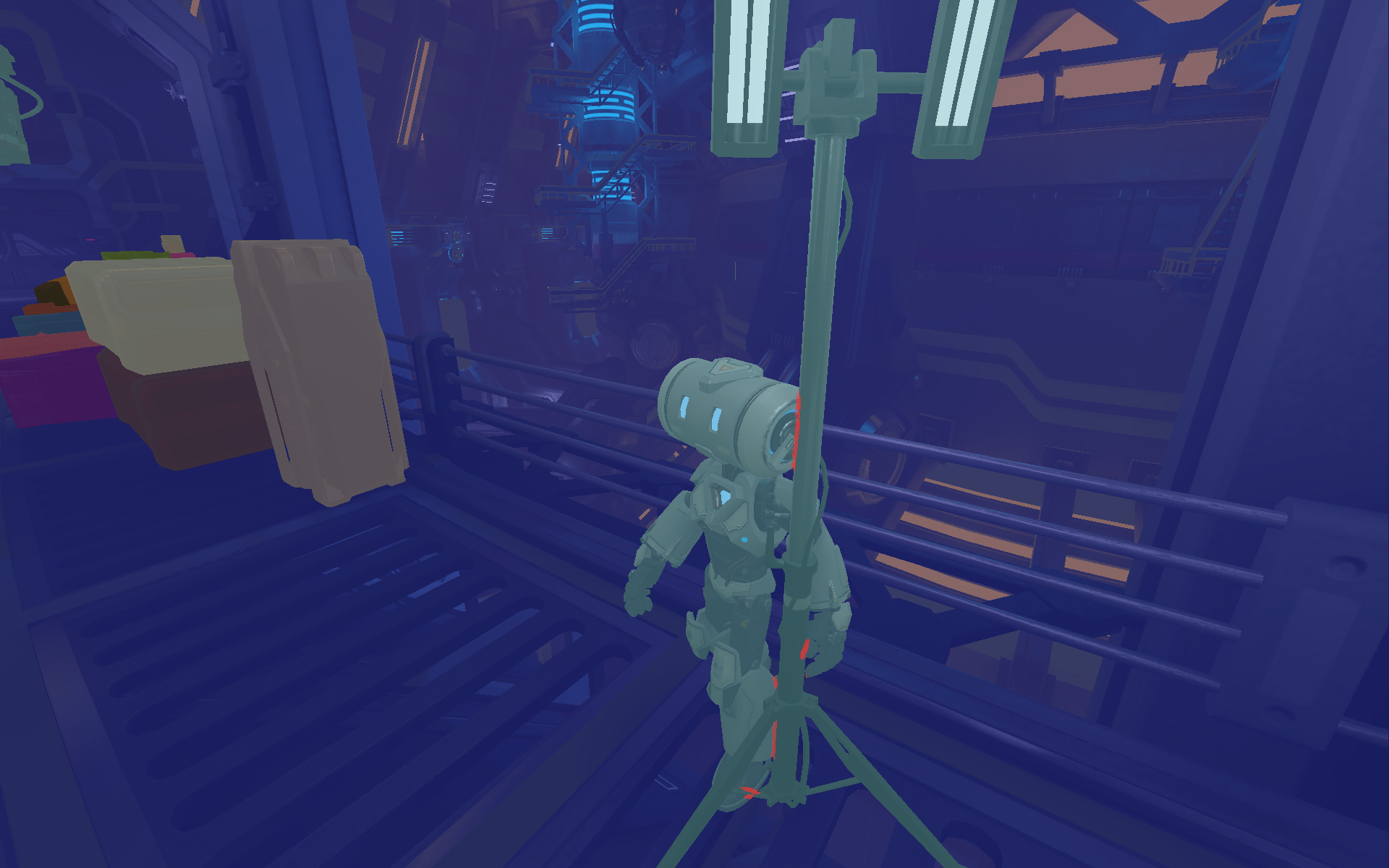}\hfill\includegraphics[width=0.495\columnwidth]{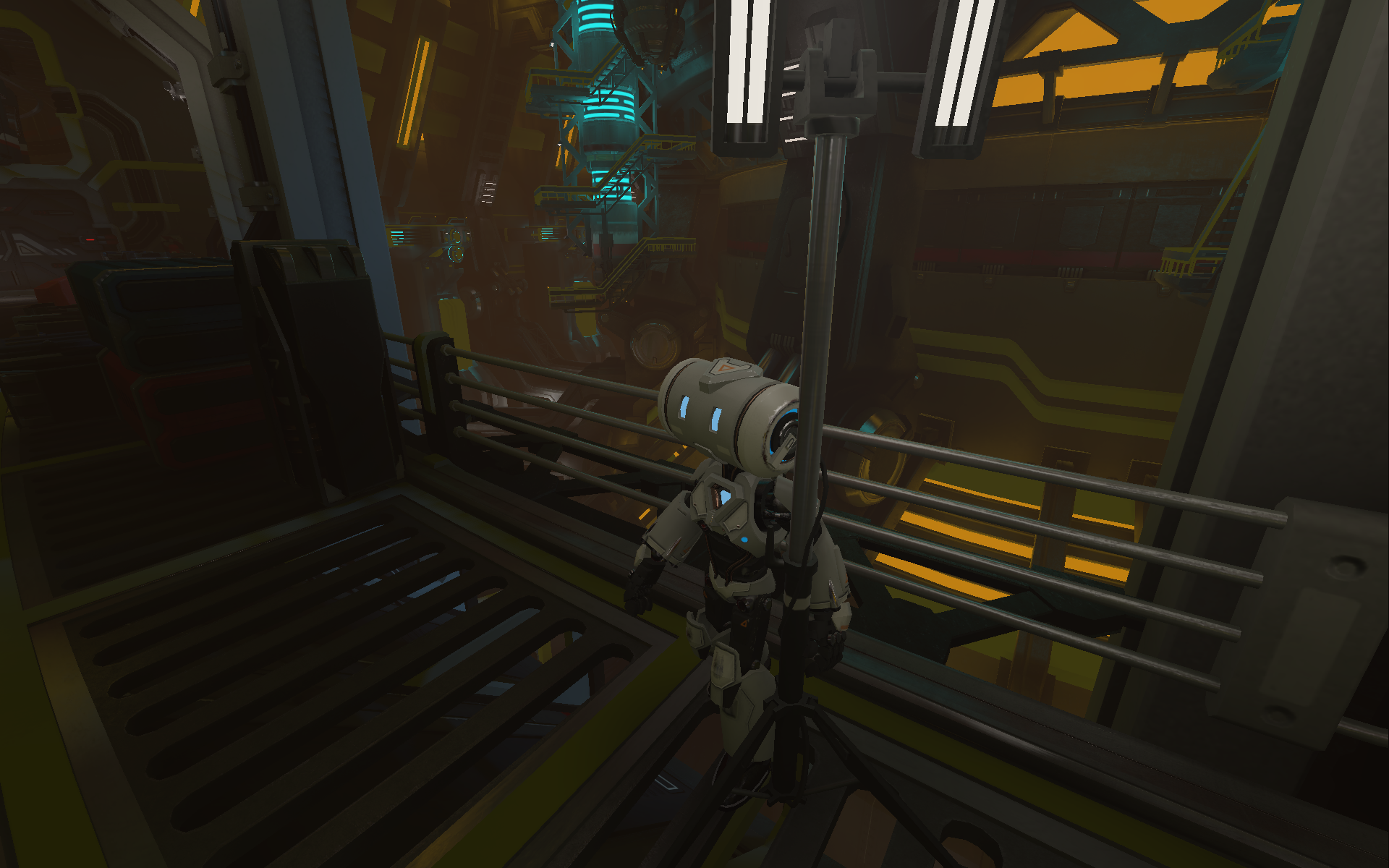}
%\caption{Hard Clipping Bug} 
\end{subfigure}

\caption{Examples of clipping bugs (right) and a visualisation of their associated object ID and clipping annotation masks overlaid (left).}
\label{fig:clipping_examples}
\end{figure}

%By combining automated environment traversal with zero-shot VLM-based analysis, we aim to bridge the gap between large-scale data generation and actionable anomaly detection, enabling fully automated testing for specific visual bugs.

The primary motivation of this work is to investigate whether automated, agent-driven anomaly detection in video games using general-purpose VLMs can serve as an effective and scalable solution for automated game QA without requiring domain-specific training. 
We focus on geometry clipping anomalies as a challenging but representative problem as shown in Fig.~\ref{fig:clipping_examples}. 
Our contributions include (i) an end-to-end pipeline that integrates autonomous exploration and VLM-based anomaly detection, and (ii) a systematic evaluation across state-of-the-art closed- and open-source VLMs for this task, including analysis of prompt sensitivity and failure modes.

\section{Experiment Setup}
\begin{figure}
    \centering
    \includegraphics[width=0.99\linewidth]{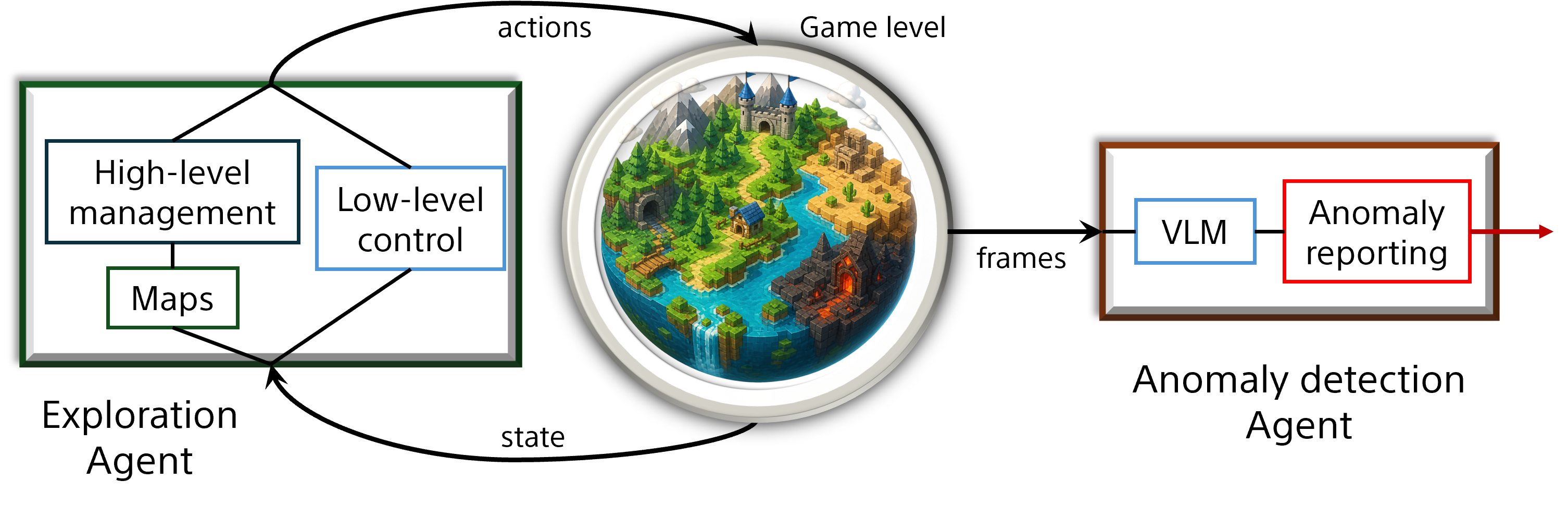}
    \caption{System architecture.}
    \label{fig:diagram}
\end{figure}

Our proposed game level testing agent is based on two main components: (i) an exploration component, which controls agent navigation to maximize coverage of the game world (Section \ref{subsec:Explore}), and (ii) a detection component based on VLMs, which processes the collected data to detect bugs or \textit{anomalies}.
These two components/agents interaction with the game level is illustrated in Fig.\ref{fig:diagram}.

The system and experiments presented in this work use open-source assets from the Godot Engine, specifically the Godot Third-Person Shooter demo project.\footnote{https://godotengine.org/asset-library/asset/678} We make minor modifications to this project to support our experiments.

\subsection{Exploration Agent}
\label{subsec:Explore}

%The agent follows an explorative strategy which uses a low-level controller that randomly selects actions (move, jump) and alters camera perspective/heading direction, and a high-level controller which determines the next region to explore, teleporting the agent to under-explored regions periodically. We follow both the assumptions and the objective functions of the exploration agents in \cite{lu2022go}. In order to maintain scalability, the strategy used to represent visitation maps is the one proposed in \cite{celemin2024bayesian}.
%
The objective of this system is to enable the agent to navigate the environment and interact with game objects while efficiently covering all reachable areas of the level.
This is a task that is difficult for human testers to perform exhaustively in a balanced manner.
This system is composed of two levels of abstraction that control the character.

\textit{Low-level control:} A stochastic open-loop policy controls the agent’s movement, enabling exploratory behavior and interaction with surrounding entities.

\textit{High-level manager:} This component determines the next region for exploration over a given time horizon, with the aim of maximizing information acquisition. 
Such a manager is inspired by \cite{lu2022go,celemin2024bayesian}.
We follow both the assumptions and the objective functions of the exploration agents in \cite{lu2022go}, which are valid during the development time of the game, but we use the map representations proposed in \cite{celemin2024bayesian} to maintain scalability.

%The explorative agent is integrated with Godot using a custom setup that is optimized for streaming image data to Python. We opted for this custom setup over the popular Godot RL package \cite{beeching2021godot} as we found that sending frame capture data was not well supported/optimized, in part also due to limitations in Godot's rendering API.

\subsection{Bug Generation \& Annotation}

Inspired by \cite{wilkins2022}, we build an automated annotation pipeline in a modified Godot 4 engine to replace manual, non-scalable labeling. 
%The modifications expose additional annotation buffers through the public rendering API, which we capture using Godot's compositor system. 
 
%This shader-based approach allows many different visual bugs to be annotated while addressing limitations of using a single binary indicator, such as whether the bug is currently visible to the player. %The setup is very general and is intended to support our on-going efforts to evaluate approaches to automated bug detection. 
% The mask is computed by checking an object ID texture and depth texture, if two texels have different object ids but the same depth then they are consider part of the intersection boundary.

\textit{Geometry Clipping:}
This common bug arises when meshes of apparently solid objects visibly intersect. Our pipeline uses shaders to produce binary masks of visible bug regions in each frame. As shown in Fig.~\ref{fig:clipping_examples}, the mask marks boundaries between intersecting objects in clipping cases.
%In practice, game developers allow certain kinds of geometry clipping issues to persist due to practical limitations. 
%For example, objects placed on uneven terrain may slightly intersect the ground because perfectly non-intersecting placement is not always feasible.
%Other cases include: parts of the same rigged/skinned mesh intersect at rigging around the skeleton joints or at other pinch points, long geometry (e.g. a spear) that are not part of the collision system to avoid hindering player movement. 
In this work, we focus on some of the more severe cases that would be considered bugs in a practical development setting. Since our shader setup detects all visible geometry-clipping boundaries, we apply additional filtering to remove subtle or spurious occurrences that would not typically be considered bugs. We compute frame-level clipping labels from the annotation mask area, applying a threshold to exclude intersections too small to be perceptible. 
% Below sentence looked out of the place - we should discuss during the meeting.
%We introduced geometry clipping by disabling colliders on certain objects in the game scene, 
% (NB) I would not write it here - not necessary and kind of shows godot in bad picture
%CC: I actually agree with all of that, I didn't want to be harsh and delete everything, I polished and reduced it to less than half of what it was, but agree all those details are not relevant
%nevertheless, the exploration agent also found pre-existing bugs in the demo project.

%To this end, we modify the Godot TPS demo project so  the player is the only dynamic object, removing enemies from the scene. 
%We then keep only the intersections between the player and the environment, including terrain and other game objects. 
%We also group any meshes that belong to the same logical object (e.g., the player character) to avoid detecting subtle self intersections. 

\subsection{Dataset} \label{subsec:dataset}

% To produce our evaluation dataset, we run the explorative agent while randomly enabling/disabled colliders for a subset of objects, sampling 6 frames per second. We further filter redundant frames using the map the agent produces, for discretized location we keep 1 normal frame, and \textit{all} bugged frames if any exist. This produces a final dataset of 2420 normal frames and 514 bugged frames. 
Although the proposed testing system is designed for exhaustive exploration and continuous anomaly detection, in this experimental setup we use the exploration agent to collect data that is later downsampled to obtain a more balanced yet diverse dataset.
For that, we run the exploration agent while randomly enabling or disabling colliders for a subset of objects and sample frames at 6 Hz. 
We then remove redundant frames using the map generated by the agent: for each discretised location, we retain one normal frame and all anomaly (clipping) frames, when present. 
This process yields a final dataset of 2420 normal frames and 516 frames with clipping bugs. 

To better analyze the performance of the VLMs for detecting glitches (clipping), we manually review the normal frames and divide them into two subsets: 500 easy examples and 500 hard examples. 
The easy subset contains normal frames that are visually unambiguous, with no apparent signs of clipping or other artifacts. 
The hard subset contains normal frames that also do not contain clipping bugs, but may appear ambiguous due to factors such as lighting, object placement, occlusion, or the robot pose, making them potentially confusing even for a human reviewer. 
In addition, to remove any selection bias, we also consider a set where we randomly select 500 normal frames from the pool of 2420 normal frames. 
%The performance results of the VLMs are presented in Table X.

% To gain additional insight into the performance of the VLMs on this task of detecting anomalies, we manually review and split the normal frames into two groups - 500 easy and 500 hard. The 500 easy examples are cases where there are no it is visually obvious to a human reviewer. The 500 hard set corresponds to the normal frames which do not contain a clipping bug but due to the lightning or placement of the objects or the robot, they might appear confusing to even a manual reviewer. 

%it is very clear to a human that there is no clipping present in the frame. The 500 hard examples are ambiguous cases where no intersection boundary is visible but there \textit{may} be clipping present, e.g. if observed from an alternative point of view or where there are significant occlusions. In addition, we also consider a split of 500 normal frames randomly sampled from the 2420 normal frames.

\begin{table*}[!htbp]
\centering
\scriptsize
\setlength{\tabcolsep}{2.5pt}
\renewcommand{\arraystretch}{1.1}
\caption{Balanced frame-level glitch detection results (\%) under different normal-frame splits and prompt variants. Recall is fixed for the generic prompt because of the same bug frames across splits.}
%Recall is fixed per model across splits since the same 500 bug frames are used throughout.}
\label{tab:balanced-split-results}
\begin{tabular*}{\textwidth}{@{\extracolsep{\fill}} l r rr rr rr rr rr rr @{}}
\hline
 & \multicolumn{7}{c}{Generic Prompt} & \multicolumn{6}{c}{Prompt Sensitivity (Random Split)} \\
\cline{2-8} \cline{9-14}
 & & \multicolumn{2}{c}{Easy} & \multicolumn{2}{c}{Hard} & \multicolumn{2}{c}{Random} & \multicolumn{2}{c}{Specific} & \multicolumn{2}{c}{Stepwise} & \multicolumn{2}{c}{Context} \\
\cline{3-4} \cline{5-6} \cline{7-8} \cline{9-10} \cline{11-12} \cline{13-14}
Model & Rec. & Acc. & Prec. & Acc. & Prec. & Acc. & Prec. & Rec. & Prec. & Rec. & Prec. & Rec. & Prec. \\
\hline
Qwen3-VL         & 65.8              & 76.0 & 82.7            & 52.3 & 51.8            & 60.7 & 59.7            & 50.4 & 69.8 & 95.6 & 50.2 & 38.6 & 61.1 \\
Gemma-4          & \textbf{86.0}     & 76.2 & 71.9            & 51.6 & 50.9            & 61.0 & 57.3            & 84.8 & 65.8 & 80.8 & 66.2 & 17.8 & \textbf{89.9} \\
Ministral-3      & 84.4              & 77.0 & 73.5            & 58.2 & 55.4            & 67.7 & 63.3            & \textbf{97.2} & 50.0 & 98.2 & 49.8 & \textbf{97.6} & 50.1 \\
Llama-4 Scout    & 68.2              & 80.8 & \textbf{91.2}   & 55.8 & 54.6            & 67.4 & 67.1            & 88.6 & 57.6 & \textbf{99.6} & 50.7 & 39.8 & 63.8 \\
Gemini-3.1 Flash & 82.0              & \textbf{85.0} & 87.2  & \textbf{60.7} & \textbf{57.5} & \textbf{71.6} & \textbf{67.9} & 91.4 & 68.3 & 92.4 & 67.2 & 94.0 & 65.6 \\
GPT-5.5          & 74.0              & 80.4 & 84.9            & 55.4 & 53.9            & 65.6 & 63.4            & 53.8 & \textbf{71.2} & 39.4 & \textbf{71.1} & 25.6 & 76.6 \\
\hline
\end{tabular*}
\end{table*}

\section{Results and Discussion}

We evaluate six VLMs spanning closed- and open-source model families. The closed-source models are Gemini-3.1-Flash-Lite-Preview and OpenAI GPT-5.5. The open-source models are Qwen3-VL-30B-A3B-Instruct-GGUF, Gemma-4-31B-it-GGUF, Llama-4-Scout-17B-16E-Instruct, and Ministral-3-14B-Instruct-2512-GGUF. The open-source models are implemented using the Unsloth framework. All models are queried in a zero-shot setting using a generic glitch-detection prompt following prior work~\cite{taesiri2025videogameqa,yakun2026eccv}. In addition, we evaluate three alternative prompt sets, described in Section~\ref{subsec:promptsens}, to ensure that the reported results are not specific to a single prompt formulation.

%All models are queried zero-shot with a generic glitch-detection prompt as used in existing literature~\cite{taesiri2025videogameqa,yakun2026eccv}; no task-specific examples are provided, neither any fine-tuning of the open-source models is performed.

\subsection{Evaluation Protocol}
\label{subsec:balanced-split-eval}

Using the three normal-frame subsets described in Section~\ref{subsec:dataset}, we construct balanced 500–500 splits by pairing each subset with the same 500 bug frames and treating the bug class as positive. To assess sensitivity to the prompt used, we evaluate four prompt variants: a \textit{generic glitch} prompt that asks the model to flag any visual anomaly; a \textit{specific clipping} prompt that narrows the task to geometry intersection; a \textit{stepwise analysis} prompt that asks the model to describe the scene before classifying; and a \textit{specific plus context} prompt that adds game-context information and examples. 

Table~\ref{tab:balanced-split-results} summarizes the balanced frame-level detection results. The left-hand columns report accuracy, precision, and recall for the generic prompt across the Easy, Hard, and Random splits. Since the same 500 bug frames are used in each split, recall is fixed for a given model; differences across splits therefore reflect changes in the normal-frame subset and the resulting number of false positives. The right-hand columns report precision and recall for the remaining prompt variants on the Random split.

\subsection{Detection Performance considering Generic Prompt}

\begin{figure}[t]
\centering
\begin{subfigure}{0.48\columnwidth}
\centering
\includegraphics[width=\linewidth]{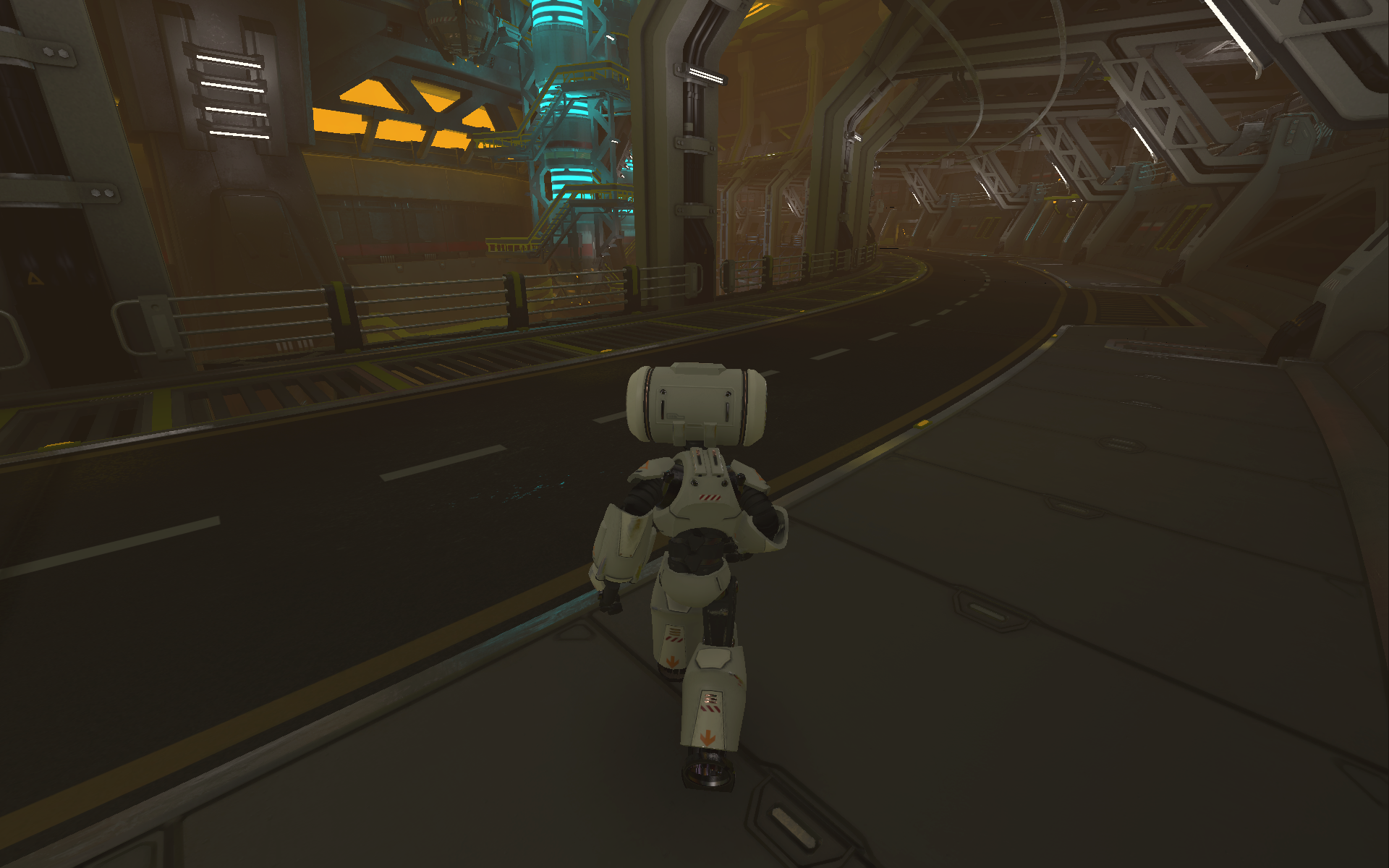}
\caption{Easy split}
\label{fig:easy-example}
\end{subfigure}\hfill
\begin{subfigure}{0.48\columnwidth}
\centering
\includegraphics[width=\linewidth]{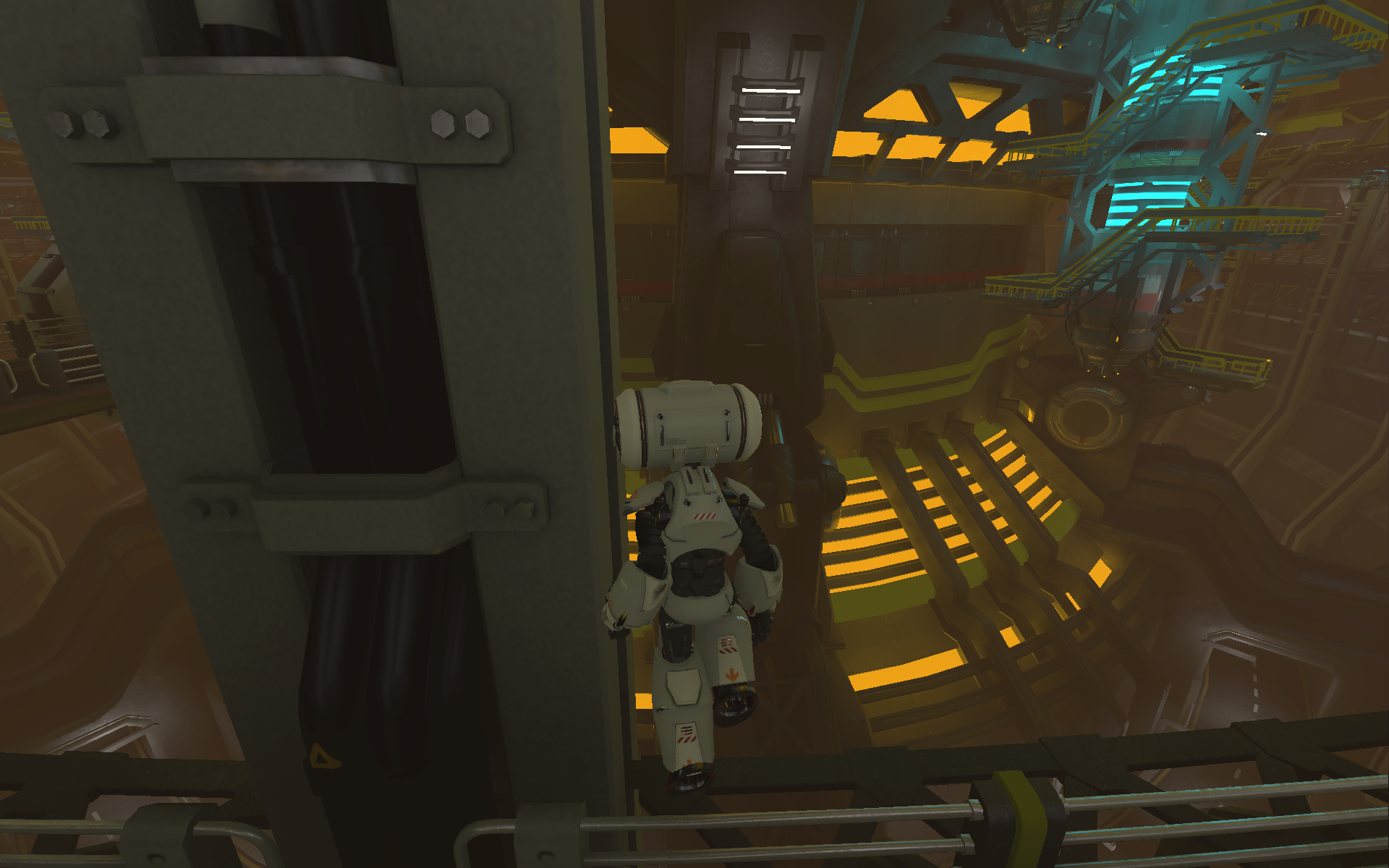}
\caption{Hard split}
\label{fig:hard-example}
\end{subfigure}
\caption{Examples of normal frames from the easy and hard splits. The easy frame (a) shows the character in an open area with no nearby geometry, clearly free of clipping. The hard frame (b) shows the character near walls and edges, creating visual ambiguity that VLMs frequently misclassify as clipping.}
\label{fig:easy-hard-examples}
\end{figure}

% On the \emph{easy} split, where negative examples are visually unambiguous (Fig.~\ref{fig:easy-example}), all models achieve their strongest results. Gemini leads with $84.5$ F1 and $85.0\%$ accuracy, while Llama attains the highest precision ($91.2\%$) at the cost of lower recall ($68.2\%$). 
% At the other end of the spectrum, Gemma and Ministral both exceed $84\%$ recall but with comparatively lower precision ($71.9\%$ and $73.5\%$, respectively). 
% These results confirm that VLMs can learn visual features associated with geometry intersection when the normal; buggy distinction is unambiguous.

% On the \emph{easy} split, where negative examples are visually unambiguous (Fig.~\ref{fig:easy-example}), all models achieve their strongest results. Gemini leads with $84.5$ F1 and $85.0\%$ accuracy, while Llama attains the highest precision ($91.2\%$) at the cost of lower recall ($68.2\%$). 
% At the other end of the spectrum, Gemma and Ministral both exceed $84\%$ recall but with comparatively lower precision ($71.9\%$ and $73.5\%$, respectively). 
% These results confirm that VLMs can reliably detect clipping when the normal frames are visually distinct from the buggy ones.

On the \emph{easy} split, where negative examples are visually unambiguous (Fig.~\ref{fig:easy-example}), all models achieve their strongest results. Gemini leads with $85.0\%$ accuracy and $87.2\%$ precision, while Llama attains the highest precision ($91.2\%$) at the cost of lower recall ($68.2\%$).
At the other end of the spectrum, Gemma and Ministral both exceed $84\%$ recall but with comparatively lower precision ($71.9\%$ and $73.5\%$, respectively).
These results confirm that VLMs can reliably detect clipping when the normal frames are visually distinct from the buggy ones.

% On the \emph{easy} split, where negative examples are visually unambiguous, all models achieve their strongest results. Gemini leads with $84.5$ F1 and $85.0\%$ accuracy, while Llama attains the highest precision ($91.2\%$) at the cost of lower recall ($68.2\%$). At the other end of the spectrum, Gemma and Ministral both exceed $84\%$ recall but with comparatively lower precision ($71.9\%$ and $73.5\%$, respectively). These results confirm that VLMs can learn visual features associated with geometry intersection when the normal--buggy distinction is unambiguous.

Performance degrades substantially on the \emph{hard} split. 
These frames contain no ground-truth clipping, yet they feature the character positioned near object boundaries, partial occlusions, or tight camera angles that visually resemble intersection geometry (Fig.~\ref{fig:hard-example}).
Every model's precision drops markedly; Gemini from $87.2\%$ to $57.5\%$, Llama from $91.2\%$ to $54.6\%$, Gemma from $71.9\%$ to $50.9\%$; while recall remains unchanged by construction. The uniform direction of this shift indicates that the additional false positives are driven by spatially ambiguous content rather than random misclassification.

% Performance degrades substantially on the \emph{hard} split. These frames contain no ground-truth clipping, yet they feature the character positioned near object boundaries, partial occlusions, or tight camera angles that visually resemble intersection geometry. Every model's precision drops markedly; Gemini from $87.2\%$ to $57.5\%$, Llama from $91.2\%$ to $54.6\%$, Gemma from $71.9\%$ to $50.9\%$; while recall remains unchanged by construction. The uniform direction of this shift indicates that the additional false positives are driven by spatially ambiguous content rather than random misclassification.

The \emph{random} split provides the most realistic operating point, reflecting the natural distribution of normal frames collected by the exploration agent. Results fall consistently between the easy and hard extremes, confirming that the hard set functions as a stress test rather than a representative sample. On this split, Gemini remains the strongest performer with an accuracy and precision of $71.6\%$ and $67.9\%$, followed by Ministral and GPT.

\subsection{Prompt Sensitivity} \label{subsec:promptsens}

Prompt sensitivity varies substantially across models. Gemini-3.1 Flash is the most robust, maintaining high recall ($82{-}94\%$) and stable precision ($66{-}68\%$) across all four prompts.
It also achieves the highest accuracy for every prompt variant, peaking at $74.5\%$ with the specific clipping prompt. 
GPT-5.5 shows the opposite trend: more specific prompts reduce recall sharply (from $74.0\%$ down to $25.6\%$ with the context-enriched prompt) while precision increases (up to $76.6\%$), suggesting the model becomes overly conservative when given detailed instructions; its accuracy remains moderate. 
Among open-source models, Gemma-4's recall stays above $80\%$ for three prompts but collapses to $17.8\%$ with the context-enriched prompt, while Ministral-3 achieves near-ceiling recall ($>97\%$) with the specific and stepwise prompts but at near-chance precision and accuracy ($\approx$50\%). 
These results suggest that prompt design is a model-specific parameter that can shift the precision--recall trade-off dramatically.

\subsection{Discussion}

\paragraph{Precision--recall trade-offs}

% The evaluated models span a useful spectrum of operating points for QA deployment. High-recall models such as Gemma ($86.0\%$) and Ministral ($84.4\%$) capture the vast majority of true bugs but generate many false alarms, particularly on ambiguous frames. Conservative models such as Llama favour precision ($91.2\%$ on easy normals) but miss a larger fraction of actual bugs ($68.2\%$ recall). Gemini offers the best compromise, achieving the highest F1 across all three splits while maintaining recall close to the high-recall open-source models and meaningfully higher precision on the hard ($57.5\%$ vs.\ $50.9–55.4\%$) and random ($67.9\%$ vs.\ $57.3–63.3\%$) splits. In practice, the preferred operating point depends on downstream cost: when human review capacity is limited, a high-precision model reduces triage burden; when the priority is to avoid shipping bugs, a high-recall model ensures broader coverage.

The evaluated models span a useful spectrum of operating points for QA deployment. High-recall models such as Gemma ($86.0\%$) and Ministral ($84.4\%$) capture the vast majority of true bugs but generate many false alarms, particularly on ambiguous frames. Conservative models such as Llama favour precision ($91.2\%$ on easy normals) but miss a larger fraction of actual bugs ($68.2\%$ recall). Gemini offers the best compromise, achieving the highest accuracy and precision across all three splits while maintaining recall close to the high-recall open-source models and meaningfully higher precision on the hard ($57.5\%$ vs.\ $50.9{-}55.4\%$) and random ($67.9\%$ vs.\ $57.3{-}63.3\%$) splits. In practice, the preferred operating point depends on downstream cost: when human review capacity is limited, a high-precision model reduces triage burden; when the priority is to avoid shipping bugs, a high-recall model ensures broader coverage.

\begin{figure}
    \centering
    \includegraphics[width=0.99\linewidth]{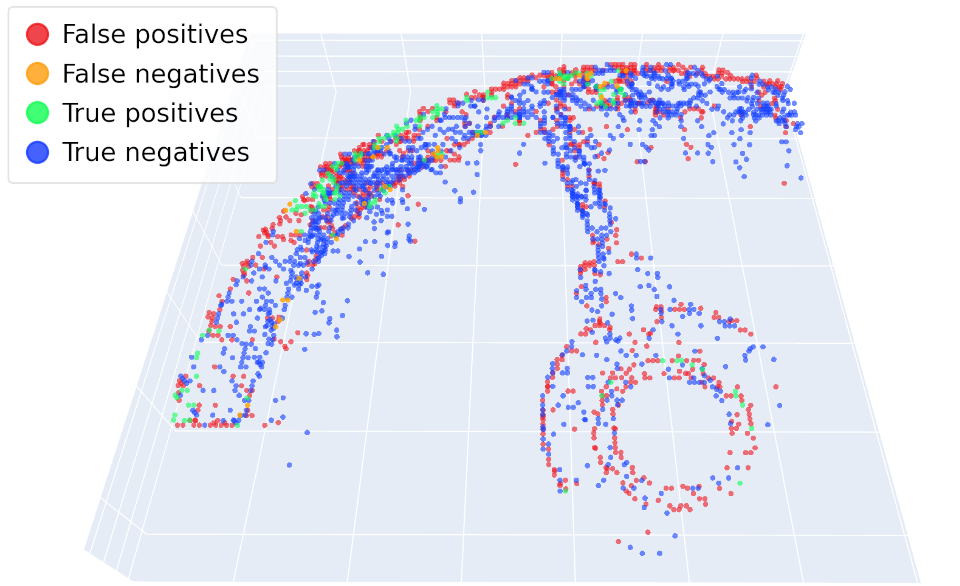}
    \caption{Agent generated map of the Godot TPS demo project level visualizing Gemini-Flash-3.1 predictions.}
    %Legend: red=false positive, green=true positive, blue=true negative, yellow=false negative.}
    \label{fig:map-example}
\end{figure}

\paragraph{Sources of false positives}
The consistent precision drop from easy to hard splits reveals a systematic failure mode shared by all evaluated models. Current VLMs appear to rely on local spatial proximity between surfaces as evidence of clipping and lack the depth reasoning needed to distinguish genuine mesh intersection from mere visual adjacency. This observation is consistent with known limitations of VLMs in fine-grained spatial reasoning~\cite{cao2024physgame, taesiri2025videogameqa}. Qualitative inspection of flagged frames supports this interpretation: many false positives involve the character standing close to walls as evidenced in Fig. \ref{fig:map-example} or being partially occluded by foreground geometry; configurations that are geometrically suspicious but do not constitute clipping bugs.

%This is further confirmed by the visualisation in \ref{fig:map-example} which shows the predictions of the Qwen3 model by spatial location. It is a trend among all the models we tested that false positives appear along the map boundaries where the player is very close/touching a wall. Additionally, we see that the use of a generic prompt leads to false positives in places where the player is in free fall, \todo{this is not seen in the other prompts - are we using them anywhere? or sticking to generic?}

\paragraph{Implications for game QA pipelines.}
These findings suggest that, in the current form, VLMs are best deployed as high-recall candidate filters within a multi-stage QA pipeline rather than as standalone bug detectors. A practical workflow would use VLM predictions to flag suspicious frames for downstream verification by human reviewers, temporal aggregation across consecutive frames, or a second-stage classifier. 
%The high recall of several models ensures broad coverage of genuine bugs, while false positives on ambiguous frames are deferred to more precise downstream steps. 
Several models ensure broad coverage of genuine bugs, while false positives on ambiguous frames are deferred to more precise downstream steps. 
The decoupled design of our pipeline, where the exploration agent handles data collection and the VLM handles initial screening, naturally supports such a staged approach: the detection back-end can be replaced or augmented without modifying the exploration strategy.

\section{Conclusions and Future Work}

% This work evaluated six VLMs for geometry clipping detection in an agent-driven game QA pipeline. Our results show that VLMs can identify visual glitches associated with geometry intersection, with the best model (Gemini-3.1 Flash) achieving $74.3$ F1 on a balanced dataset. However, all models produce substantial false positives on visually ambiguous scenes suggesting that current VLMs lack the fine-grained spatial reasoning needed for reliable game anomaly detection. These findings position VLMs as promising high-recall candidate filters within multi-stage QA workflows rather than as final bug arbiters.

This work evaluated six VLMs for geometry clipping detection in an agent-driven game QA pipeline. Our results show that VLMs can identify glitches associated with obvious geometry intersection. However, all models exhibit substantial misclassification rates on visually ambiguous frames. 
%While some errors may stem from the lack of specialized context or domain knowledge in VLMs, many cases are inherently difficult to classify, even for human testers, who would likely exhibit similarly high error rates (see challenging examples in Fig. \ref{fig:clipping_examples} and Fig. \ref{fig:hard-example}). 
The lower performance may partly reflect limitations in task-specific context or domain knowledge of existing VLMs, but many examples are also intrinsically difficult to judge from a single frame, even for human testers (see challenging examples in Fig. \ref{fig:clipping_examples} and Fig. \ref{fig:hard-example}). A future evaluation should therefore consider not only ground truth labels but also human performance, as human testers are the ultimate users this technology is intended to support.
%All models produce substantial false positives on visually ambiguous scenes suggesting that current VLMs lack the fine-grained spatial reasoning needed for reliable game anomaly detection. 

A key limitation of this evaluation is its reliance on single-frame annotations. In practice, clipping bugs often have a spatio-temporal nature; they may persist across multiple frames, emerge during character or camera motion, or depend on player interactions. While a single frame may suffice to identify obvious clipping artifacts, more subtle or transient cases may require temporal context, and VLM performance evaluation considering multiple-frames may differ from the single-frame results reported here. Moreover, our benchmark covers only one bug type in a single game environment, generalisation to other anomaly types and visual styles remains to be investigated.

The metrics reported here should be read as a characterisation of VLM capabilities rather than deployment-ready performance estimates. Future work will extend the benchmark to additional bug categories, explore intelligent frame sampling and video-based reasoning, and investigate alternative detection back-ends.

\bibliographystyle{IEEEtran}
\bibliography{references}

% \begingroup
% % \setstretch{0.94}
% \setlength\bibitemsep{0pt}
% \printbibliography
% \endgroup

\end{document}